\documentclass[lettersize,journal]{IEEEtran}
\usepackage{subcaption}
\usepackage{booktabs}
\usepackage{amsmath,amsfonts}
\usepackage{amssymb}
\usepackage{algorithm}
\usepackage{algpseudocode}
\usepackage{array}
\usepackage{textcomp}
\usepackage{multirow}
\usepackage{stfloats}
\usepackage{url}
\usepackage{verbatim}
\usepackage{graphicx}
\usepackage{hyperref}
\usepackage{enumitem}
\usepackage{cite}
\algnewcommand\Input{\item[\textbf{Input:}]} 
\algnewcommand\Output{\item[\textbf{Output:}]} 
\algrenewcommand\alglinenumber[1]{\scriptsize #1:}
\newtheorem{definition}{Definition}

\begin{document}

\title{Approximate Subgraph Matching with Neural Graph Representations and Reinforcement Learning}

\author{IEEE Publication Technology,~\IEEEmembership{Staff,~IEEE,}
\author{Kaiyang Li, Shihao Ji,~\IEEEmembership{Senior Member,~IEEE}, Zhipeng Cai,~\IEEEmembership{Fellow,~IEEE}, and Wei Li%
\thanks{Kaiyang Li and Shihao Ji are with the University of Connecticut, 352 Mansfield Road, Storrs, CT 06269, USA (e-mail: kaiyang.li@uconn.edu; shihao.ji@uconn.edu).}%
\thanks{Zhipeng Cai and Wei Li are with Georgia State University, 33 Gilmer Street SE, Atlanta, GA 30303, USA (e-mail: zcai@gsu.edu; wli28@gsu.edu).}}

\thanks{This paper was produced by the IEEE Publication Technology Group. They are in Piscataway, NJ.}
\thanks{Manuscript received April 19, 2021; revised August 16, 2021.}}

\markboth{Journal of \LaTeX\ Class Files,~Vol.~14, No.~8, August~2021}%
{Shell \MakeLowercase{\textit{et al.}}: A Sample Article Using IEEEtran.cls for IEEE Journals}


\maketitle

\begin{abstract}
Approximate subgraph matching (ASM) is a task that determines the \emph{approximate} presence of a given query graph in a large target graph. Being an NP-hard problem, ASM is critical in graph analysis with a myriad of applications ranging from database systems and network science to biochemistry and privacy. Existing techniques often employ heuristic search strategies, which cannot fully utilize the graph information, leading to sub-optimal solutions. This paper proposes a Reinforcement Learning based Approximate Subgraph Matching (RL-ASM) algorithm that exploits graph transformers to effectively extract graph representations and RL-based policies for ASM. Our model is built upon the branch-and-bound algorithm that selects one pair of nodes from the two input graphs at a time for potential matches. Instead of using heuristics, we exploit a Graph Transformer architecture to extract feature representations that encode the full graph information. To enhance the training of the RL policy, we use supervised signals to guide our agent in an imitation learning stage. Subsequently, the policy is fine-tuned with the Proximal Policy Optimization (PPO) that optimizes the accumulative long-term rewards over episodes. Extensive experiments on both synthetic and real-world datasets demonstrate that our RL-ASM outperforms existing methods in terms of effectiveness and efficiency. Our source code is available at \url{https://github.com/KaiyangLi1992/RL-ASM}.
\end{abstract}


\section{Introduction}
Graph analysis, a sub-field of data mining, has gained popularity in recent years as graph structured data becomes increasingly ubiquitous in a broad range of domains~\cite{aggarwal2010managing,chakrabarti2022graph}. One of the fundamental problems of graph analysis is subgraph matching, which identifies the presence of a query graph in a large target graph. Subgraph matching has a wide variety of applications, including database retrieval~\cite{10.14778/2002974.2002976}, knowledge graph mining~\cite{8085196}, biomedical analysis~\cite{tian2007saga}, social group finding~\cite{10.1145/2274576.2274578}, and privacy protection~\cite{lu2024neural}. 

In most of the subgraph matching studies, the basic assumption is that graphs are noise-free and accurate. In order to identify an occurrence of a query graph, all nodes and edges of the query graph must occur in the target graph. In other words, the matching has to be exact. However, noise commonly exists in many real-world subgraph matching scenarios. Therefore, the Approximate Subgraph Matching (ASM) has emerged as a more practical task in graph analysis. For example, noise is prevalent in protein-protein interaction (PPI) networks~\cite{zaslavskiy2009global} due to errors in data collection and varying experimental thresholds. By discovering and analyzing approximate matches, biologists can still validate their hypotheses based on noisy protein interaction network data. Another example is social network de-anonymization~\cite{ji2016graph}, which involves identifying anonymous individuals within a social network by correlating nodes from an auxiliary network with those from a target network, where the auxiliary network is typically a noisy subgraph. Effective matching between these two graphs enables adversaries to launch de-anonymization attacks. Therefore, research into ASM is important to evaluate data vulnerability and develop privacy-preserving solutions. 
\begin{figure}[!t]
\centering
  \includegraphics[width=0.4\textwidth]{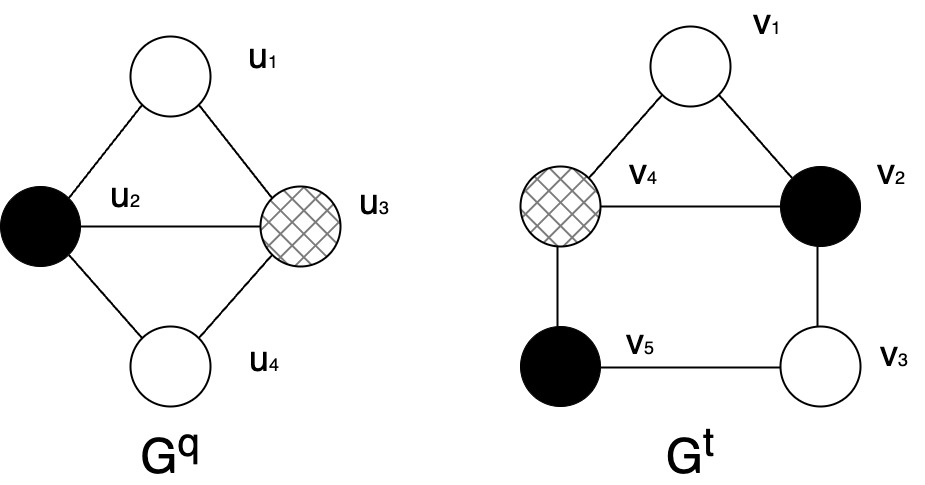}\vspace{-8pt}
  \caption{An example of approximate subgraph matching, where $\{(u_1, v_1), (u_2, v_2),$ $(u_3, v_4), (u_4, v_3)\}$ achieves the best approximate matching from query graph $G^q$ to target graph $G^t$ with the smallest graph edit distance of 1.}
  \label{fig:example}\vspace{-10pt}
\end{figure}

Formally, given a query graph $G^q$, ASM aims to find a subgraph in target graph $G^t$ that has the smallest graph edit distance (GED) to $G^q$. An example is illustrated in Fig.~\ref{fig:example}, where the mapping $\{(u_1, v_1), (u_2, v_2),$ $(u_3, v_4), (u_4, v_3)\}$ achieves the best approximate match from $G^q$ to $G^t$ with the smallest GED of 1. The existing ASM algorithms can be mainly categorized into two classes: (1) the methods that convert ASM to exact subgraph matching~\cite{reza2020approximate,zhang2010sapper}, and (2) the methods that employ a branch-and-bound algorithm for combinatorial optimization~\cite{tu2020inexact}. The first category of methods specifies a threshold $k$ and reduces the ASM problem to identifying all exact matches between any prototype graphs and target graph, where the prototype graphs are all graphs that can be derived from the query graph with the GED smaller than $k$. Therefore, the exact matches of the prototype graphs are the approximate matches of the query graph. However, these methods only consider prototypes by adding edges to query graphs, excluding those that are derived by altering nodes' labels. This limitation renders these methods unsuitable for scenarios where nodes' labels are noisy. One of the reasons why these methods exclude prototype graphs generated by altering nodes' labels is that doing so would significantly increase the number of prototype graphs. For example, consider a graph $G^q$, which is a cycle graph $C_8$, where each node has 16 possible label categories. With a GED threshold $k=3$, considering only edge-induced distances, it results in 1,351 prototype graphs. However, if both edges and node labels are considered in GED, this number escalates to 347,971.

On the other hand, the branch-and-bound based ASM methods match one pair of nodes from query graph and target graph one at a time and extend intermediate results iteratively. For example, Tu et al.~\cite{tu2020inexact} calculate the GED lower bounds of the best possible solutions within each branch of the search tree, then compare these lower bounds with the GED of the current best match, and prune the branches that are not possible to yield an optimal solution, i.e., the branches with lower bounds greater than the GED of the current best match. However, these methods select mapping node pairs with a greedy strategy based on heuristics, i.e., selecting the node pair that leads to the branch with the minimized GED lower bound at each step by utilizing nodes' local structural and label information. Therefore, let alone its greedy nature, the method lacks the ability of fully utilizing the graph information in the two input graphs, leading to sub-optimal solutions.

To address these limitations of the existing approaches, we introduce a Reinforcement Learning based Approximate Subgraph Matching (RL-ASM) method. RL-ASM employs a Graph Transformer~\cite{rampavsek2022recipe} to extract feature representations from graphs, whose performances in graph representation learning have been extensively validated both theoretically and empirically, allowing the use of full graph information for ASM. To mitigate the issues induced by greedy algorithms, we train our neural network based agent with RL algorithms to optimize an accumulative long-term reward over episodes. As a result, our RL-ASM achieves the solutions that are closer to the optimal ones, outperforming the existing search algorithms by a significant margin. The contributions of our work are summarized as follows:
\begin{itemize}[leftmargin=*]
\item To the best of our knowledge, RL-ASM is the first work that leverages reinforcement learning  for high-quality approximate subgraph matching.
\item Our RL-based Graph Transformer model can fully utilize the graph information and select the node pairs by optimizing an long-term reward instead of being greedy.
\item Extensive experiments conducted on synthetic and real-world graph datasets demonstrate that RL-ASM outperforms existing methods in terms of effectiveness and efficiency.
\end{itemize}

\section{Related Work}
\textbf{Traditional ASM Methods.} Being an NP-hard problem~\cite{garey1979computers}, ASM has been tackled in various approaches. Tong et al.~\cite{tong2007fast} investigate ASM in social networks and propose a method that samples the matched subgraph via random walk. Tian et al.~\cite{tian2007saga} and Yuan et al.~\cite{yuan2015efficient} break the query graph into small fragments and assemble the matches of these fragments to generate the entire match of the query graph. Other works~\cite{dutta2017neighbor, agarwal2024venom, agarwal2020chisel} utilize the chi-square statistics to measure the node similarity based on the label distribution of the node's neighbors and apply the similarity measure to search for ASM. Sussman et al.~\cite{sussman2019matched} propose to calculate the permutation matrix on adjacency matrix and match query graph to target graph via the permutation matrix. Recently,  Reza et al.~\cite{reza2020approximate} and Zhang et al.~\cite{zhang2010sapper} identify all prototype graphs whose distances from query graph are below a specified threshold, then reduce the ASM problem to the exact subgraph matching problem by searching for exact matches between the prototype graphs and target graphs. Tu et al.~\cite{tu2020inexact} approach the ASM problem by formulating it as a tree search problem, employing lower bounds and cutoffs to reduce the search space. Given a sufficient running time, this method can find an optimal solution, i.e., the subgraph of target graph that is the most similar to query graph.

Unfortunately, most of the existing methods utilize hand-crafted features to search for the mapped node pairs, and cannot fully exploit the graph information for matching. What's more, they all exploit greedy algorithms that overlook the potential better matches in future steps, leading to sub-optimal solutions.


\vspace{5pt}
\noindent\textbf{Reinforcement Learning for NP-hard Graph Problems.}
There are many prior works focusing on RL algorithms to solve NP-hard problems on graphs, including Minimum Vertex Cover~\cite{khalil2017learning}, Network Dismantling~\cite{fan2020finding}, and Maximum Common Subgraph detection~\cite{liu2020learning,bai2021glsearch}. As of exact subgraph matching, Wang et al.~\cite{wang2022reinforcement} utilize RL to determine the node mapping order on query graphs, while Bai et al.~\cite{bai2022detecting} utilize RL to determine the node mapping order on target graphs. Compared with exact subgraph matching, ASM represents an even more challenging task since ASM allows discrepancies between query graph $G^q$ and the matched subgragh of $G^t$. Therefore, hard constraints that typically can be used to reduce the search space (e.g., requiring mapped nodes to have the same label) are not applicable. To address this, our RL-ASM employs a branch-and-bound algorithm to reduce the search space and leverages a Graph Transformer to extract graph features and select actions from a large action space.

\section{Preliminary}
\subsection{Problem Definition}

Let $G = (V,E)$ denote a graph with a set of nodes $V$ connected by a set of edges $E$. The approximate subgraph matching problem can be defined as follows.

\begin{definition}
\textbf{Approximate Subgraph Matching (ASM):} Given a query graph $G^q = (V^q, E^q)$ and a target graph $G^t = (V^t, E^t)$, ASM aims to identify a one-to-one mapping ${M: V^q \rightarrow V^t}$ such that the graph edit distance $C(M; G^q, G^t)$ is minimized, where $C(M; G^q, G^t)$ measures the discrepancies between query graph $G^q$ and the subgraph of $G^t$ induced by $M(V^q)$.
\end{definition} 

Following~\cite{tu2020inexact,he2006closure}, we adopt the graph edit distance $C(M; G^q, G^t)$ over nodes and edges as
\begin{equation} 
C(M; G^q, G^t)\!=\!\!\sum_{u \in V^q }\! D_{V}(u, M(u)) +\! \sum_{e \in E^q }\! D_{E}(e, M(e)),
\label{eqn:dist} 
\end{equation} 
where the node distance is
\begin{equation} 
D_{V}(u, M(u)) = \begin{cases} 0 & \mathcal{A}(u)=\mathcal{A}(M(u)) \\  1  & \mathcal{A}(u) \neq \mathcal{A}(M(u))  \end{cases}, 
\label{eqn:dist_node} 
\end{equation} 
and the edge distance is  
\begin{equation} 
D_{E}(e, M(e)) = \begin{cases} 0 & e \in E^q, M(e) \in E^t \\  1  &  e \in E^q, M(e) \notin E^t  \end{cases}.
\label{eqn:dist_node} 
\end{equation} 
Here $\mathcal{A}(u)$ denotes the label of node $u$. Let $e$ be the edge between nodes $u_1$ and $u_2$ from $G^q$, and $M(e)$ refers to the edge between $M(u_1)$ and $M(u_2)$ from $G^t$.

\begin{figure*}
  \includegraphics[width=\textwidth]{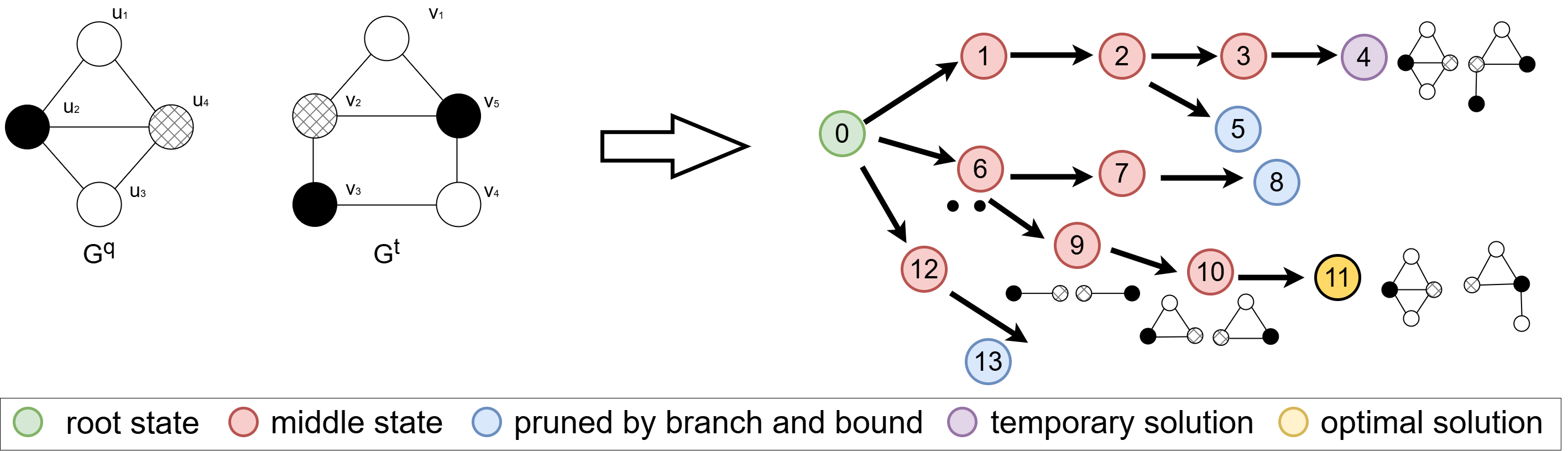}\vspace{-1pt}
  \caption{An illustration of the search process of ASM on $(G^q, G^t)$.  The branch-and-bound search algorithm (Algorithm~\ref{alg:1}) produces a tree structure, where each node represents a state ($s_t$), the node ID reflects the order in which the state is visited, and each direct edge represents an action ($a_t$), which adds a node-pair to the current node-node mapping $M_t$. The search is essentially depth-first with pruning through the lower bound check. The policy (Line 12 of Algorithm~\ref{alg:1}) refers to a node-pair selection strategy, i.e., which state to visit next? The visiting order affects the performance of searching in the tree. For example, if state 6 can be visited before state 1, a better solution can be found in fewer iterations. This means the GED of the current best match $currentMiniDist$ will be smaller, allowing the algorithm to prune more branches in subsequent search steps. Hence, the search efficiency will be higher. When the search completes or a pre-defined search iteration budget is exhausted, the best solution identified by then will be returned. For the clarity of visualization, some nodes and edges in the search tree are omitted.} 
  \label{fig:SearchTree}\vspace{-5pt}
\end{figure*}

\subsection{The Search Algorithm for ASM}
Since the branch-and-bound search algorithm serves as the foundation of our proposed method, we first review a typical branch-and-bound search algorithm for ASM~\cite{tu2020inexact}. We then discuss its limitations and constraints, which lead to our proposed RL-ASM.

As shown in Algorithm~\ref{alg:1} and Fig.~\ref{fig:SearchTree}, the branch-and-bound search algorithm, as proposed in~\cite{tu2020inexact}, starts from an initial state with an empty mapping $M_0$, and adds one pair of nodes to the mapping each time, while maintaining the best solution found so far. At each search step, denote the current search state as $s_t$, which consists of $G^q$, $G^t$ and the current node-node mapping $M_t$. The algorithm attempts to select a node pair $(u, v)$, where $u\in G^q$ and $v\in G^t$, as action $a_t$, and add the pair to $M_t$. As shown in Fig.~\ref{fig:SearchTree}, each action (represented by a directed edge in the search tree) updates one state to another. For example, action ($u_2$, $v_5$) updates state 0 to state 6, starting a branch that includes states 7-11. In~\cite{tu2020inexact}, the authors propose a method to calculate the lower bound of $C(M;G^q,G^t)$ for action selection. At Line 7 of Algorithm~\ref{alg:1}, the algorithm calculates the lower bound for each branch resulting by an action from action space $A_t$. For example, as shown in Fig.~\ref{fig:SearchTree}, when the current state is state 0, the algorithm calculates the lower bounds corresponding to the branches starting from states 1, 6, and 12. Then the algorithm compares these lower bounds with $currentMiniDist$ -- the  GED of the current best match. Any actions leading to the branches with the lower bounds greater than $currentMiniDist$ will be eliminated from action space $A_t$ (Line 8). If all the actions in $A_t$ are eliminated, the algorithm will backtrack to the parent search state (Lines 9-11), i.e., the current branch will be cut off. When all of the nodes in $G^q$ have been mapped, the algorithm compares its $C(M_t;G^q,G^t)$ with $currentMiniDist$. If $C(M_t;G^q,G^t)$ is smaller, indicating a better solution than current best mapping is identified, the algorithm sets the $C(M_t;G^q,G^t)$ as the new  $currentMiniDist$ and updates $M_t$ to $M^*$ (Lines 17-20). Note that if the method can find good matches early, $currentMiniDist$ will be small, which means a lot of branches would be pruned and the search space could be reduced substantially. 


\begin{algorithm}
\caption{Branch-and-Bound for Approximate Subgraph Matching~\cite{tu2020inexact}}
\label{alg:1}
\begin{algorithmic}[1]
\Input  query graph $G^q$ and  target graph $G^t$
\Output optimal node-node mapping $M^*$
\State Initalize $s_0$  $\gets$  ($G^q$, $G^t$, $M_0=\varnothing$)
\State Initalize $stack$ $\gets$ new Stack($s_0$)
\State Initalize $currentMiniDist$ $\gets$ $\infty$ 
\While{$stack$ $\neq$ $\varnothing$}
\State $s_t$ $\gets$ $stack$.pop()
\State $A_t$ = $s_t$.get\_actionspace()
\State $Lbounds$  $\gets$ $s_t$.get\_lowerbounds($A_t$)
\State $A_t$  $\gets$ prune\_actionspace($A_t$,$Lbounds$,$currMiniDist$)
\If{  $|A_t|$ = 0}
\State continue
\EndIf
\State $a_t$ $\gets$ policy($s_t$, $A_t$)
\State $A_t$ $\gets$ $A_t - \{a_t\}$
\State $M_t$  $\gets$ $s_t$.get\_mapping()
\State $M_t$  $\gets$ $M_t$  + $a_t$
\If{ $|M_t|$  = $|V^q|$ }
\If{ $C(M_t; G^q, G^t) < currMiniDist$}
\State $currMiniDist$ $\gets$ $C(M_t; G^q, G^t)$
\State $M^*$ $\gets$ $M_t$
\EndIf
\State continue
\EndIf
\State $stack$.push($s_t$)
\State $s_{t+1}$ $\gets$ environment.update($s_t$, $a_t$) 
\State $stack$.push($s_{t+1}$)
\EndWhile
\end{algorithmic}
\end{algorithm}

The above method leverages a heuristic for action selection, denoted as ``policy"  at Line 12. Typically, a greedy policy is adopted, where the action leading to the branch with the minimum lower bound of GED is selected. A significant limitation of this method is that the lower bound based heuristics is not adaptive to the complex real-world graph structures because the method cannot fully exploit the information embedded in the graphs. More importantly, the greedy algorithm focuses on a locally optimal objective for action selection at the current step without considering potential better matches in the future steps, leading to sub-optimal solutions.

\section{Proposed method} \label{sec:methods}
In this section, we introduce our Reinforcement Learning-based Approximate Subgraph Matching (RL-ASM). We provide a high-level overview of RL-ASM in Section~\ref{sec:overview}, including the definitions of state, action, and reward. Section~\ref{sec:framework} describes the details of our framework, focusing on the design of representation learning with Graph Transformer for ASM. This is followed by Sections~\ref{sec:ppo} and~\ref{sec:pre-training}, where the model training is elaborated. In Section~\ref{sec:explain}, we discuss why structural encoding features and Graph Transformer are required for ASM. 

\subsection{Overview}~\label{sec:overview}
We formulate the ASM as a Markov Decision Process (MDP). The state $s_t$ consists of two components: (1) query graph $G^q$ and target graph $G^t$, and (2) the current node-node mapping $M_t$ from $V^q$ to $V^t$. Action $a_t: u\to v$ is defined as adding a node pair $(u, v)$ to $M_t$, where $u\in V^q$ and $v\in V^t$. Our policy assigns a score to each action in action space $A_t$, and is modeled as a neural network $P_\theta(a_t|s_t)$, parameterized by $\theta$, that computes a probability distribution over $A_t$ given the current state $s_t$.

For subgraph matching, the immediate reward $r_t(s_t,a_t)=r_t^{\text{node}}+r_t^{\text{edge}}$ consists of two components: the node-matching reward $r_t^{\text{node}}$ and the edge-matching reward $r_t^{\text{edge}}$. The node-matching reward $r^\text{node}_t$ is determined by the label compatibility of the nodes added by $a_t: u\to v$, yielding +1 for identical labels and -1 otherwise. 
The edge-matching reward $r^\text{edge}_t $ is defined as $|E_q^+|-|E_q^-|$.  
Here, let $E_q$ be the existing edges between node $u$ and all the mapped nodes of $G^q$ at state $s_t$. $E_q^+$ consists of all edges $e\in E_q$ whose mapped edge $M_t(e)$ exists in $E^t$. Conversely, \(E^-_q\) includes those edges $e\in E_q$ whose mapped edge does not exit in $E^t$. For instance, at state 10 of Fig.~\ref{fig:SearchTree} with $M_{10}=\{(u_1, v_1), (u_2, v_5), (u_4, v_2)\}$ and $a_{10}: u_3 \to v_4$, $E^+_q=\{(u_2, u_3)\}$ since \( (v_4, v_5) \), the mapped edge, is in \( E^t \), while $E^-_q=\{(u_3, u_4)\}$ since \( (v_2, v_4) \), the mapped edge, is not in \( E^t \). Thus, $r_{10}^{\text{edge}}=0$. Essentially, $r_t(s_t,a_t)$ measures the change of GED after adding node pair $(u,v)$ (a.k.a taking action $a_t$) to the current mapped subgraphs of query graph and target graph at state $s_t$.

\begin{figure*}
  \includegraphics[width=\textwidth]{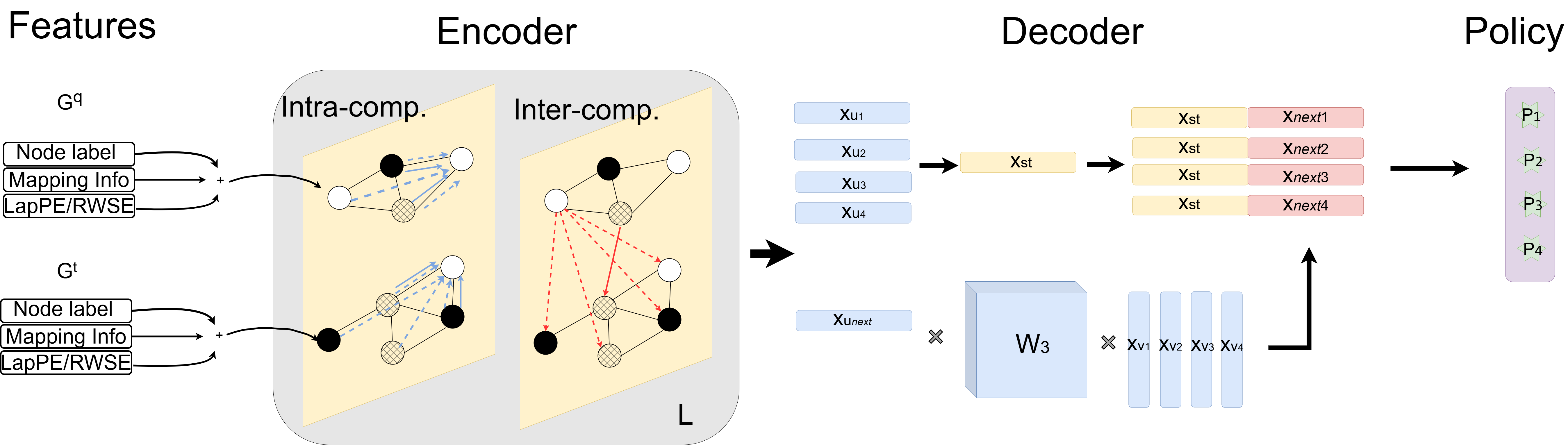}
  \caption{Overview of RL-ASM. RL-ASM consists of two major components: encoder and decoder. The encoder processes node label, mapping info, positional and/or structural encodings by alternating intra- and inter-components $L$ times (with $L$ layers) to extract powerful node representations. The decoder leverages self-attention to node embeddings of $G^q$ to generate a global state representation $\mathbf{x_{s_t}}$. The action representations are then generated by the product of embeddings of node $u_{next}\in G^q$ (according to $\phi$), learnable weight tensors $\mathbf{W_3}$, and embeddings of unmapped candidate nodes $v_1,v_2,v_3$, and $v_4$ from $G^t$. Subsequently, the representations of state and actions are concatenated, which is fed to a MLP and a softmax classifier to calculate the probability distribution over the actions $P_{\theta}(a_t|s_t)$.}
  \label{fig:framework}
\end{figure*}

If we do not impose any constrains to the action space, the cardinality of the action space will be $O(|V^q|\times|V^t|)$, which can be prohibitively huge. To reduce the action space size, we generate $\phi$, the mapping order of nodes in $G^q$, before searching for subgraph matches. At each step, we take one node from $G^q$ according to the order $\phi$, and map this node to one of the nodes in $G^t$. We generate the order $\phi$ with the method proposed in~\cite{bonnici2013subgraph}, which is designed for the exact subgraph matching~\cite{sun2020memory}. The algorithm starts by identifying the node $u\in V^q$ that possesses the maximum degree, and adds it to $\phi$. In the subsequent step, the algorithm prefers to add those nodes that have a larger number of edges with the nodes already present in $\phi$. This is an effective method to create $\phi$ because the node with the highest degree and the ones with more mapped neighbors typically encode substantial structural information, and thereby facilitate accurate pattern matching to initiate the search.

To enhance the efficiency in the tree search, we further design a mechanism that caches the lower bounds of branches and the policy scores of actions evaluated in previous steps. This allows our method to reuse cached results when backtracking to earlier states. Because of the strong correlation between the lower bounds and the GED of current best solution $currMinDist$~\cite{tu2020inexact}, we delete the cached results whenever $curMinDist$ is changed, and recalculate the lower bounds and the policy scores when backtracking to these states.

\subsection{The RL-ASM Framework}~\label{sec:framework}
As shown in Fig.~\ref{fig:framework}, our RL-ASM framework consists of an encoder that produces the node embeddings for $V^q$ and $V^t$, and a decoder that transforms the embeddings into a probabilistic distribution over actions $P_\theta(a_t|s_t)$ for action selection. To design this model, we face several challenges:
(i) many existing graph neural networks (GNNs), which primarily follow a message passing paradigm, are inherently local and their expressiveness cannot exceed the 1-Weisfeiler-Lehman test~\cite{xu2018how}. Hence, these models lack the capability to tackle the complex approximate subgraph matching problem (see Section~\ref{sec:explain} for justifications);
(ii) as the task involves both query and target graphs, effectively sharing information between the two graphs presents a significant challenge; and (iii) each state $s_t$ contains $G^q$, $G^t$, and the node mapping $M_t$; it is challenging to effectively encapsulate the entire state's information into a representation that can be utilized to predict $P_\theta(a_t|s_t)$. These challenges have guided the design of our model.

\subsubsection{Node Features}
We employ three distinct features for each node:
(1) \textbf{Label encoding}: A one-hot vector that specifies the label of each node;
(2) \textbf{Mapping status}: A binary number indicating the node's mapping status - 1 if mapped, and 0 otherwise; this feature is useful to alleviate Challenge (iii);
(3) \textbf{Positional and structural encodings}: For node's positional information, we utilize the Laplacian Positional Encoding (LapPE), which involves the eigenvectors associated with the $k$ smallest non-zero eigenvalues of the graph Laplacian~\cite{NEURIPS2021b4fd1d2c}. For structural information, we apply Random Walk Structural Encoding (RWSE), derived from the diagonal of the $m$-step random walk matrix~\cite{dwivedi2021graph}. Rampášek et al.~\cite{rampavsek2022recipe} and Kreuzer et al.~\cite{NEURIPS2021b4fd1d2c} have demonstrated theoretically and empirically that using positional encoding and structural encoding in GNNs can enhance the model's expressiveness. Depending on the characteristics of the graph, we choose either positional or structural encoding or both as node features. This alleviates Challenge (i). Positional encoding is particularly effective for image graphs~\cite{han2022vision}, while structural encoding is well suited for molecular graphs~\cite{sun2022does}. Each of the three features is processed by a distinct linear layer and subsequently concatenated as the feature representations, which are fed to the encoder.

\subsubsection{Encoder}
The encoder consists of an intra-component module and an inter-component module as shown in Fig.~\ref{fig:framework}. The intra-component module processes and aggregates messages within query graph $G^q$ and target graph $G^t$, separately. The inter-component module shares messages across the graphs based on the relationships of nodes in $G^q$ and $G^t$.

To alleviate Challenge (i) and effectively utilize both local and global graph information, we employ the GraphGPS layer~\cite{rampavsek2022recipe} as the intra-component. This hybrid layer combines a Message Passing Neural Network (MPNN) and a global attention mechanism, and offers greater expressiveness than traditional MPNNs, such as GCN~\cite{kipf2016semi}, GAT~\cite{velivckovic2017graph}, and GraphSAGE~\cite{xu2018how}. The MPNN propagates messages along the edges, whereas the global attention spreads information throughout the entire graph, as illustrated by the solid and dotted blue lines in Fig.~\ref{fig:framework}. The global attention mechanism enhances this capability by passing messages across all nodes, and therefore overcomes the expressiveness bottleneck associated with over-smoothing and over-squashing~\cite{rampavsek2022recipe,akansha2023over}. In each GraphGPS layer, node features are updated by integrating outputs from both the MPNN and the global attention instances. Specifically, the GraphGPS layer is formulated as 
\begin{align}
&\mathbf{X^{\ell+1}_{intra}} = \text{GPS}^\ell(\mathbf{X^\ell},  \mathbf{A}),  \\
\intertext{which is implemented as}
&\mathbf{X_M^{\ell+1}} = \text{MPNN}^\ell(\mathbf{X^\ell}, \mathbf{A}),  \\
&\mathbf{X_H^{\ell+1}} = \text{GlobalAttn}^\ell(\mathbf{X^\ell}),  \\
&\mathbf{X^{\ell+1}_{intra}} = \text{MLP}^\ell(\mathbf{X_M^{\ell+1}} + \mathbf{X_H^{\ell+1}}), 
\end{align}
where \(\mathbf{A} \in \mathbb{R}^{N \times N}\) is the adjacency matrix of a graph of \(N\) nodes; \(\mathbf{X^\ell} \in \mathbb{R}^{N \times d}\) denotes the \(d\)-dimensional node features for $N$ nodes at the \(\ell\)-th layer; \(\text{MLP}^\ell\) is a 2-layer multilayer perceptron (MLP) block; and \(\mathbf{X^{\ell+1}_{intra}} \in \mathbb{R}^{N \times d}\) represents the output of the intra-component at the \(\ell\)-th layer. \(\text{MPNN}^\ell\) and \(\text{GlobalAttn}^\ell\) are modular components, which are implemented as GatedGCN~\cite{li2016gated} and Self-Attention~\cite{vaswani2017attention}, respectively.

As for the inter-component module of the encoder, we employ a cross-attention mechanism that enables the model to learn how to share messages between $G^q$ and $G^t$. This mechanism is established based on the node mapping relationships. For state \( s_t \), the current mapping $M_t$ is $\{u \rightarrow v \mid u \in V^q_t, v \in V^t\}$, where \( V^q_t \) represents the set of nodes of \( G^q \) that have been mapped at step $t$. Our method allows the nodes of $G^q$ to be mapped to nodes in $G^t$ even if their labels differ. Consequently, for any node $u'\in G^q$ that remains unmapped, we define its candidate set \( C_{u'} \) as $ \{ v' \in V^t \mid v' \text{ is unmapped in } G^t\}$. Therefore, the mapping describing the candidate relationships between nodes can be formulated as $M'_t = \{u' \rightarrow C_{u'}, \forall u' \in V_q \setminus V^q_t\}$. Let $\widetilde{M_t}$ denote the union of $M_t$ and $M'_t$, and $ \widetilde{M_t}^{-1}$ denote the reverse mapping of $\widetilde{M_t}$. The messages passing along $M_t$ and $M'_t$ are shown as solid and dotted red lines in Fig.~\ref{fig:framework}, respectively.  Specifically, the message passing from $G^q$ to $G^t$ is implemented as follows:
\begin{equation}
\mathbf{X^\ell_{inter}}[u,:] \!=\! \mathbf{W}_1^\ell \mathbf{X^\ell_{intra}}[u,:] +\!\!\!\! \sum_{v \in \widetilde{M_t}(u)}\!\!\!\! \alpha_{uv} \mathbf{W}_2^\ell \mathbf{X^\ell_{intra}}[v,:],
\label{eq:cross_attention}
\end{equation}
where $\mathbf{X^\ell_{inter}}[u,:]$ and $\mathbf{X^\ell_{intra}}[u,:]\in\mathbb{R}^d$ represent inter- and intra-embeddings of node $u\in V_q$ at the $\ell$-th layer, respectively; the attention coefficients $\alpha_{uv}$ are derived using multi-head dot-product attention~\cite{vaswani2017attention}, and  $\mathbf{W}_1^\ell$ and $\mathbf{W}_2^\ell$ are the learnable matrices associated with the $\ell$-th layer. Additionally, we implement the message passing from $G^t$ to $G^q$ following the reverse mapping $\widetilde{M_t}^{-1}$. The inter-component module is crucial for information exchange across the graphs, which addresses Challenge (ii).

The output from the inter-component module is then fed to an activation function, denoted as $\mathbf{X^{\ell+1}} = f_{activate}(\mathbf{X^{\ell+1}_{inter}})$, yielding the output of the $\ell$-th encoder layer. We construct the encoder by stacking four such layers and employ  jump knowledge~\cite{xu2018representation} to aggregate the outputs from these layers.

\vspace{10pt}
\subsubsection{Decoder}
The node embeddings of query graph $\mathbf{X^q} \in \mathbb{R}^{|V^q|\times d}$, produced by the encoder, capture the information from $G^q$ and from $G^t$ through the cross-graph message passing manifested by $\widetilde{M_t}$ and $\widetilde{M_t}^{-1}$. This allows $\mathbf{X^q}$ to gather comprehensive information required to represent the entire state. To alleviate Challenge (iii), we introduce an attention-based mechanism that calculates the state embedding $\mathbf{x_{s_t}}$ based on the embeddings of the query graph nodes as follows:
\begin{equation}
\mathbf{x_{s_t}} = \text{Pooling}(\text{SelfAttention}(\mathbf{X^q})).
\end{equation}
Here, we utilize a self-attention mechanism~\cite{shaw2018self} on  $\mathbf{X^q}$, followed by aggregation of the resulting representations using a pooling operation. In our experiments, average pooling is employed for aggregation.

Given that the policy is formulated as \( P_{\theta}(a_t|s_t) \), it is essential to learn the representations of state \( s_t \) and action \( a_t \). Consider action $a_t: u \to v$, with $\mathbf{x}_u$\footnote{Here, $u$ is the next node selected from $G^q$ according to the order $\phi$.} and $\mathbf{x}_v \in \mathbb{R}^d$ denoting the node representations learned by the encoder, we employ a bilinear tensor product defined by a learnable tensor $\mathbf{W}_3 \in \mathbb{R}^{d \times d \times F}$ to ensure sufficient  interaction between the node embeddings involved by action. Here, $F$ is a hyperparameter that enhances the model's capacity by adding dimensions to learn complex relationships. Then action $a_t: u \to v$ can be expressed as $\mathbf{x}_u^T \mathbf{W}_3 \mathbf{x}_v$, which is concatenated with the state embedding $\mathbf{x}_{s_t}$. The combined vector is then fed into a multi-layer perceptron (MLP), followed by a softmax layer to generate 
\begin{equation}
\!\!P_{\theta}(a_t | s_t)\!=\!\text{SoftMax}(\text{MLP}(\text{CONCAT}(\mathbf{x}_{s_t}, \mathbf{x}_u^T \mathbf{W}_3 \mathbf{x}_v))).
\end{equation}

\subsection{Policy Training}~\label{sec:ppo}
The goal of policy training is to maximize the accumulative long-term reward: $R_t = \sum_{i=t}^{T} \gamma^{i-t} r_i$, where $T$ is the total number of time steps of an episode, $\gamma$ is the discount factor, and $r_i$ is the immediate reward incurred by action $a_i$ at the $i$-th step. To enhance training efficiency, we disable the backtracking mechanism during the policy training. That is, once our algorithm reaches to a leaf node in the search tree, the episode ends. We employ the Proximal Policy Optimization (PPO)~\cite{schulman2017proximal}, one of the most effective policy gradient methods, to train our model. PPO mitigates excessively large policy changes by clipping policy probability ratios, which helps maintain stability in the learning process. The core term of the PPO loss function is defined as
\begin{equation}
L^{clip}(\theta)\!=\!-\hat{\mathbb{E}}_t\!\!\left[\min\! \left(\!\rho(\theta)\!\hat{R}_t, \operatorname{clip}\!\left(\rho(\theta),\! 1\!-\!\varepsilon,\! 1\!+\!\varepsilon\right) \!\hat{R}_t\!\right)\!\right],
\end{equation}         
where $\hat{\mathbb{E}}_t[\cdot]$ indicates the empirical average over a batch of samples, \( \rho(\theta) = \frac{P_{\theta}(a_t|s_t)}{P_{\theta_{old}}(a_t|s_t)} \) is the ratio of the action probabilities under current and old policy, and $\hat{R}_t$ denotes the accumulated reward, normalized across the batch. 
The operation $\operatorname{clip}(\rho(\theta), 1-\varepsilon, 1+\varepsilon) \hat{R}_t$ modifies the objective by clipping the probability ratio, thereby moderating the influence of extreme policy changes during the training. To ensure sufficient exploration and prevent premature convergence to a local optimal policy, we also incorporate an entropy bonus term to the loss function:
\begin{equation}
L^{entrop}(\theta) = -\hat{\mathbb{E}}_t[H(P_{\theta}(\cdot|s_t))],
\end{equation}
where $H(\cdot)$ is the entropy of the probability distribution over action space $A_t$ given state $s_t$.  
The overall objective function of PPO is
\begin{equation}
L^{PPO}(\theta) = L^{clip}(\theta) +  cL^{entrop}(\theta),  
\end{equation}  
where $c$ is a hyperparameter that balances the contributions from the two loss terms. 

\subsection{Pre-training}~\label{sec:pre-training}
For graph pairs of large graphs, the action space can be huge. 
To expedite the learning process and enhance sample efficiency, we pre-train our model with imitation learning before the PPO fine-tuning. This pre-training involves sampling subgraphs from the target graph and recording the correspondence between nodes of the sampled and target graphs to define expert actions. Noise is added to these sampled graphs to generate query graphs.
In this pre-training stage, the model's predictions are compared to the expert’s choices, with  training  conducted using a supervised cross-entropy loss:
\begin{equation}
L^{Imit}(\theta) = -\sum_{a \in A_t} y_a \log (\hat{y}_a(\theta)),
\end{equation} 
where \( y_a \) indicates whether the action \( a \) aligns with the expert's choice (1 for yes, 0 for no), and \( \hat{y}_a \) represents the predicted probability corresponding to action \( a \). Note that during imitation learning, we enable Dropout and Batch Normalization in the model, whereas in the PPO fine-tuning, we disable these components to improve the training stability.


\subsection{Why are LapPE and Graph Transformer Needed for ASM?}~\label{sec:explain}
\begin{figure}
  \includegraphics[width=0.48\textwidth]{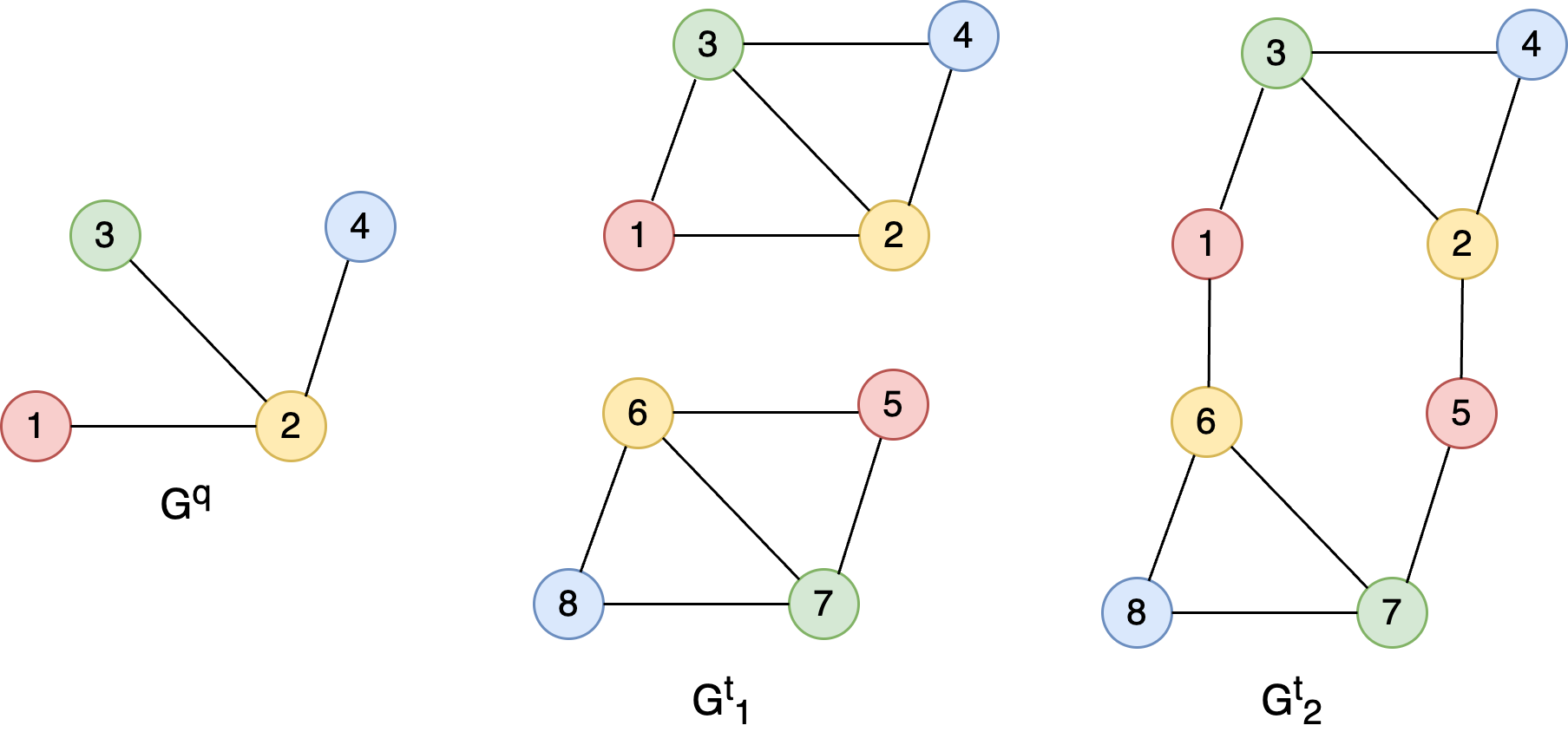}\vspace{-5pt}
  \caption{Example graphs that are indistinguishable by MPNN.}
  \label{fig:appendex}
\end{figure}

The encoder of RL-ASM consists of an intra-component module and an inter-component module.  In the inter-component, messages are passed along $\widetilde{M}$ and $\widetilde{M}^{-1}$, which denote the edges between the nodes in $G^q$ and $G^t$. This implies that the inter-component can be considered as an MPNN.
Therefore, if we employ traditional MPNN architectures (e.g., GCN, GAT, or GatedGCN) as the intra-component of our encoder, the entire architecture will be an MPNN. It has been proven that the expressiveness of MPNNs cannot exceed the 1-Weisfeiler-Lehman  test~\cite{xu2018how}. That is, they cannot differentiate non-isomorphic graphs by using only 1-hop neighborhood aggregation. In Fig.~\ref{fig:appendex}, when only the labels of nodes (which correspond to node colors in the figure) are used as input features, the MPNN fails to distinguish between $G^t_1$ and $G^t_2$ because the nodes with the identical ID in $G^t_1$ and $G^t_2$ have the same 1-hop neighborhood structure. Consequently, when attempting to match query graph $G^q$ with $G^t_1$ and $G^t_2$, an MPNN-based model will yield identical matching results. However, the optimal matching solutions for these cases are different; one is \{(1,1), (2,2), (3,3), (4,4)\} and another one is \{(1,5), (2,2), (3,3), (4,4)\}. This example illustrates the expressiveness limitation of MPNNs when addressing the ASM problem.

Kreuzer et al.~\cite{NEURIPS2021b4fd1d2c} demonstrate that with the full set of Laplacian eigenvectors, a Graph Transformer model can be a universal function approximator on graphs and can provide an approximate solution to the graph isomorphism problem. 
This suggests that by integrating LapPE and GraphGPS (a type of Graph Transformer) into the model, our RL-ASM has the capability beyond the traditional MPNN models in solving the ASM problem.

\section{Experiments}
In this section, we compare our RL-ASM with state-of-the-art approaches for ASM, and demonstrate its effectiveness and efficiency. All our experiments were conducted using a server equipped with Intel Xeon Gold 6242 CPUs and NVIDIA Tesla V100 GPUs. For reproducible research, we provide our source code at \url{https://github.com/KaiyangLi1992/RL-ASM}.

\subsection{Datasets}
Our experiments utilize one synthetic dataset and three real-world datasets from diverse domains: biology, computer vision, and social networks. These datasets are described as follows:

\begin{itemize}[leftmargin=*]
    \item \textbf{SYNTHETIC}~\cite{Morris+2020}: This dataset comprises 300 synthetic graphs generated by a statistical model. Each graph consists of 100 nodes and 196 edges, and each node is endowed with a normally distributed scalar label.

    \item \textbf{AIDS}~\cite{Morris+2020}: This dataset contains 2,000 graphs, each representing a molecular compound. Nodes in these graphs correspond to atoms, while edges represent the chemical bonds between them.

    \item \textbf{MSRC\_21}~\cite{Morris+2020}: As a semantic image processing benchmark, this dataset includes 563 graphs, each representing the graphical model of an image. Nodes in the graph represent objects in an image, and edges illustrate the relationships among the objects.

    \item \textbf{EMAIL}~\cite{leskovec2007graph}: Originating from an email communication network at a European research institution, this dataset uses nodes to represent employees, and edges to represent the email interactions between employees. The node label indicates the department to which each employee belongs.
\end{itemize}

For datasets \textbf{SYNTHETIC}, \textbf{AIDS}, and \textbf{MSRC\_21}, which contain multiple graphs, we randomly select graphs from these collections to serve as the target graphs. For the \textbf{EMAIL} dataset, which contains a single graph, we use this graph consistently as the target graph. We employ the Breadth-First Search (BFS) on these target graphs to sample nodes, ceasing the sampling when the node count reaches the parameters specified in Table~\ref{tab:dataset}. From these sampled nodes, induced subgraphs are generated to form the \emph{seed} query graphs. We then introduce noise to these \emph{seed} query graphs by adding edges and altering node labels, thereby producing query graphs with noise levels of 0\%, 5\%, and 10\%. Specifically, we calculate the total number of nodes and edges in each query graph, multiply this total number by the noise level to determine the noise intensity, \(N_{noise}\). We then randomly select an integer \(N^{node}_{noise}\) between 0 and \(N_{noise}\), change the labels of \(N^{node}_{noise}\) nodes, and add \(N_{noise} - N^{node}_{noise}\) edges to the original query graph, thereby generating a noisy query graph.
Each pair of a query graph and its corresponding original target graph constitutes a graph pair for ASM. After eliminating duplicate graph pairs, we partition the dataset into training, validation, and test sets following an 8:1:1 ratio. The statistics of the query and target graphs are reported in Table~\ref{tab:dataset}.

\begin{table}[t]
\centering
\begin{tabular}{@{}l@{}cccc@{}}
\hline
Dataset & \textbf{SYNTHETIC} & \textbf{AIDS} & \textbf{MSRC\_21} & \textbf{Email} \\ 
\hline
Target size (nodes) & 100 & 17.03 & 77.48 & 1005  \\
Target size (edges) & 196 & 17.58 & 198.18 & 25571 \\
Query size (nodes) & 15.75  & 8.09 & 24.12 & 13.19  \\
Query size (edges) & 18.13 & 8.24 & 57.29 & 60.04 \\
Num. of Node Labels & 13 & 40 & 27 & 47 \\
Sampling range of $|V^q|$ & [10, 20] & [6, 10] & [16, 32] & [8, 16] \\
\hline
\end{tabular}
\caption{Statistics of query graphs and target graphs.}
\label{tab:dataset}\vspace{-10pt}
\end{table}

To closely simulate real-world conditions, we operate under the assumption that the noise level in query graphs remains unknown. To ensure our model adapts broadly,  our training and validation sets for each dataset encompass query graphs from all three noise levels: 0\%, 5\%, and 10\%. However, for the purpose of evaluation, we specifically test our model's performances on these noise levels to report its performances under different conditions. 

\begin{table*}[t]
\centering
\begin{tabular}{cccccc}
\hline
\textbf{Training Stage}&\textbf{Hyperparameter} & \textbf{AIDS} & \textbf{SYNTHETIC} & \textbf{MSRC\_21} & \textbf{EMAIL} \\
\hline
\multirow{4}{*}{\rotatebox[origin=c]{90}{\textbf{}}}
& Hidden Dimension  & 32 & 32 & 32 & 32 \\
&\# Encoder Layers & 4 & 4 & 4 & 4 \\
& Interaction Dimension F & 32 & 32 & 32 & 32 \\
&Positional and Structural Encodings & RWSE & RWSE & LapPE & RWSE \\
\hline
\multirow{4}{*}{\rotatebox[origin=c]{0}{\textbf{Imitation Learning}}}
&Batch Size & 1024 & 1024 & 1024 & 1024 \\
&Learning Rate & 0.001 &0.0005 & 0.001 & 0.001 \\
&\#Epochs & 1000 & 1000 & 1000 & 1000 \\
& Weight Decay & 0.01 & 0.01 & 0.01 & 0.01 \\
\hline
\multirow{8}{*}{\rotatebox[origin=c]{0}{\textbf{PPO Fine-tuning}}}
&Learning Rate &0.005	&0.001	&0.0005	&0.0005 \\
&Entropy Coefficient & 0.01 & 0.01 & 0.01 & 0.01 \\
&Batch Size  &2048	&512	&1024	&64 \\
&\# Epochs & 10 & 10 & 10 & 10 \\
& $\gamma$ & 0.99 & 0.99 & 0.99 & 0.99 \\
&Clipping Range & 0.2 & 0.2 & 0.2 & 0.2 \\
\hline
\end{tabular}
\caption{The hyperparameters of RL-ASM used for different benchmark datasets.}
\label{tab:hyperparams}
\end{table*}

\subsection{Baselines}
We compare our method with a neural-network based method NeuroMatch~\cite{lou2020neural} and two heuristic methods ISM~\cite{tu2020inexact} and APM~\cite{reza2020approximate}. As shown in Fig.~\ref{fig:SearchTree},  a good initial  matching solution is crucial for the  search performance of branch-and-bound methods. Therefore, we present the first-round matching results (i.e., the results without backtracking) for both ISM and our method in this section. Additionally, we emphasize the role of imitation learning as an important pre-training stage of our method, showcasing the performances of our model when trained exclusively through imitation learning. We benchmark against the following methods:

\begin{itemize}[leftmargin=*]
    \item \textbf{NeuroMatch}~\cite{lou2020neural} decomposes query and target graphs into small subgraphs and embeds them using graph neural networks. It first predicts the probability of nodes in query graphs matching to nodes in target graphs with the embeddings of the decomposed subgraphs. It then utilizes the Hungarian algorithm~\cite{kuhn1955hungarian} to map the nodes between the two graphs such that the sum of the probabilities of the matched node pairs is maximized with the constraint that each node in query graph is assigned to a different node in target graph.

    \item \textbf{APM}~\cite{reza2020approximate} reduces ASM to exact subgraph matching. It focuses on identifying exact matches between prototype graphs (i.e., those derived from the query graph with GED below a specific threshold) and target graphs. To avoid an excessively large search space, the method only considers prototypes that are derived by adding edges to query graphs, excluding those that are derived by altering nodes' labels. 

    \item \textbf{ISM}~\cite{tu2020inexact} formulates ASM as a tree search problem. It calculates the lower bound of the GED between query graph and the corresponding subgraph in target graph within each search branch from the current step. The branches, whose lower bounds surpass the GED of the best match found so far, are pruned. The method employs a greedy strategy to select the node pair that leads to the branch with the minimal lower bound at each search step.
   
    \item \textbf{RL-ASM-IL} showcases the first-round mapping results of our RL-ASM, which is trained exclusively through imitation learning. This illustrates the impact of imitation learning to our model.
    
    \item \textbf{RL-ASM-FR} and \textbf{ISM-FR} illustrate the first-round matching results of RL-ASM and ISM without backtracking.
\end{itemize}

\vspace{5pt}
For the experiments of ISM and RL-ASM, we limit the search tree exploration to 600 seconds. If the programs fail to traverse the entire tree within this time limit, we terminate the execution and treat the best matching result found so far as the final outcome.

The hyperparameters of RL-ASM used for different benchmark datasets are provided in Table~\ref{tab:hyperparams}. We select the best model according to the metric on the validation set and report the metric on the test set. Specifically, the learning rates for imitation learning and PPO fine-tuning are selected from \{5e-5, 1e-4, 5e-4, 1e-3, 5e-3, 1e-2\}. We search for the hidden dimension of nodes from \{16, 32, 64\} for RL-ASM.

\subsection{Results}
\subsubsection{Effectiveness of Approximate Subgraph Matching}
The graph edit distance (GED), as reported in Table~\ref{tab:results}, quantifies the discrepancies between the query graph and its corresponding subgraph in the target graph, with smaller values indicating better matches. The results of RL-ASM-IL indicate that even when trained solely via imitation learning, our model's initial matching results significantly outperform the heuristic methods: APM and ISM. This demonstrates our model’s ability to effectively learn graph information via Graph Transformer for ASM. Compared with NeuroMatch and RL-ASM-IL, RL-ASM-FR shows superior performance on the \textbf{SYNTHETIC}, \textbf{MSRC\_21}, and \textbf{EMAIL} datasets. This improvement attributes to the RL training phase, which maximizes an accumulative long-term reward. The enhancements observed from RL training on the \textbf{MSRC\_21} and \textbf{EMAIL} datasets are more pronounced compared to the \textbf{AIDS} dataset. This is because the graphs in \textbf{MSRC\_21} and \textbf{EMAIL} are much larger and more challenging, requiring the model to engage in optimizing long-term rewards and explore a broader solution space, the areas where RL excels. RL-ASM delivers the best results by effectively searching the trees with backtracking to find better subgraph matches beyond the results of RL-ASM-FR. In the following subsection, we will demonstrate that our model, guided by a neural network, can potentially identify optimal solutions within a constrained time limit.

\begin{table*}[h]
\centering
\small
\caption{Graph Edit Distance (GED) between query graph and its matched subgraph identified from target graph. Note that the ``GroundTruth" number denotes the GED between the ``noisy" query graph and its corresponding \emph{seed} query graph sampled from target graph.}
\label{tab:results}\vspace{-5pt}
\begin{tabular}{@{}lcccccccccccc@{}}
\toprule
\textbf{Dataset} & \multicolumn{3}{c}{\textbf{SYNTHETIC}} & \multicolumn{3}{c}{\textbf{AIDS}} & \multicolumn{3}{c}{\textbf{MSRC\_21}} & \multicolumn{3}{c}{\textbf{EMAIL}} \\
\cmidrule(r{1em}){1-1}\cmidrule(rl{1em}){2-4} \cmidrule(l){5-7} \cmidrule(l){8-10} \cmidrule(l){11-13}
\textbf{Noise Ratio} & \textbf{0\%} & \textbf{5\%} & \textbf{10\%} & \textbf{0\%} & \textbf{5\%} & \textbf{10\%} & \textbf{0\%} & \textbf{5\%} & \textbf{10\%}& \textbf{0\%} & \textbf{5\%} & \textbf{10\%} \\
\midrule
GroundTruth  & 0.00 & 1.70 & 3.39 & 0.00 & 1.00 & \textbf{1.67} & 0.00 & 3.99 & \textbf{8.03} & 0.00  & 3.66 & 7.32 \\
\cmidrule{1-13}
NeuroMatch~\cite{lou2020neural}            & 1.52 & 4.42 & 7.96 & 0.80 & 3.12 & 7.49 & 3.56 & 11.46 & 22.43 & 3.71  & 17.41 & 26.51 \\
APM~\cite{reza2020approximate}            & 0.00 & 2.45 & 4.57 & 0.00 & 1.24 & 2.43 & 0.00 & 7.46 & 13.43 & 0.00  & 8.11 & 15.61 \\
\cmidrule{1-13}
ISM-FR       & 8.14 & 9.13 & 11.03 & 0.79 & 1.91 & 2.92 & 18.50 & 21.66 & 26.87 & 4.97 & 11.25 & 13.32 \\
ISM~\cite{tu2020inexact}           & 7.00 & 8.35 & 10.60 & 0.01 & 1.04 & 2.00 & 17.07 & 20.69 & 26.26 & 4.67 & 10.74 & 12.90 \\
\cmidrule{1-13}
RL-ASM-IL    & 0.00 & 1.65 & 3.40 & 0.06 & 1.16 & 1.85 & 0.01 & 4.29 & 8.93 & 0.00 & 3.65 & 7.00 \\
RL-ASM-FR    & 0.00 & 1.59 & 3.31 & 0.06 & 1.17 & 1.84 & 0.00 & 4.19 & 8.93 & 0.00 & 3.56 & 6.63 \\
RL-ASM (ours) & \textbf{0.00} & \textbf{1.54} & \textbf{3.18} & \textbf{0.00} & \textbf{1.04} & 1.77 & \textbf{0.00} & \textbf{4.07} & 8.77 & \textbf{0.00} & \textbf{3.13} & \textbf{6.09} \\
\bottomrule
\end{tabular}
\end{table*}

When no noise is added to query graphs (e.g., noise ratio $0\%$), APM can find the optimal solution because it reduces the ASM to an exact subgraph matching problem. For each graph pair, it attempts to match $G^q$ and $G^t$ using an exact subgraph mapping method, which can consistently identify the optimal solution when query graph $G^q$ contains no noise. However, the performance of APM declines when handling noisy instances of $G^q$ as it does not allow mappings between node pairs with differing labels.

For ISM, which selects actions based on heuristics, it struggles to find effective solutions in the initial round, as manifested by the results of ISM-FR in Table~\ref{tab:results}. This limitation hinders its ability to prune numerous branches during tree searches. Moreover, the greedy search utilizes heuristics to make decision at each step. Consequently, both its initial and final mappings are prone to sub-optimal. In addition, NeuroMatch is designed for exact subgraph matching; when noise is added to the query graph, it struggles to map nodes accurately based on the learned representations.

\begin{figure*}[htb] 
    \centering
    \begin{subfigure}{0.24\textwidth}
        \includegraphics[width=\linewidth]{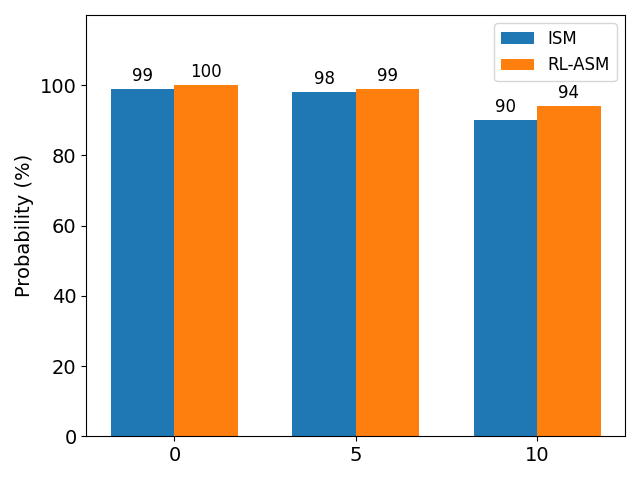} 
        \caption{AIDS}
    \end{subfigure}\hfill 
    \begin{subfigure}{0.24\textwidth}
        \includegraphics[width=\linewidth]{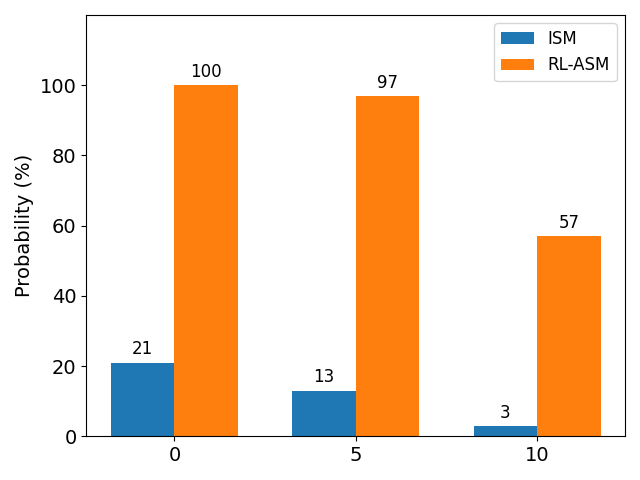} 
        \caption{SYNTHETIC}
    \end{subfigure}\hfill
    \begin{subfigure}{0.24\textwidth}
        \includegraphics[width=\linewidth]{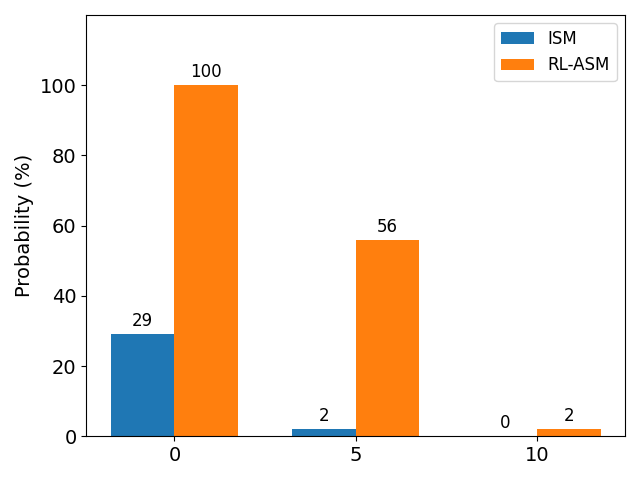} 
        \caption{MSRC\_21}
    \end{subfigure}\hfill
    \begin{subfigure}{0.24\textwidth}
        \includegraphics[width=\linewidth]{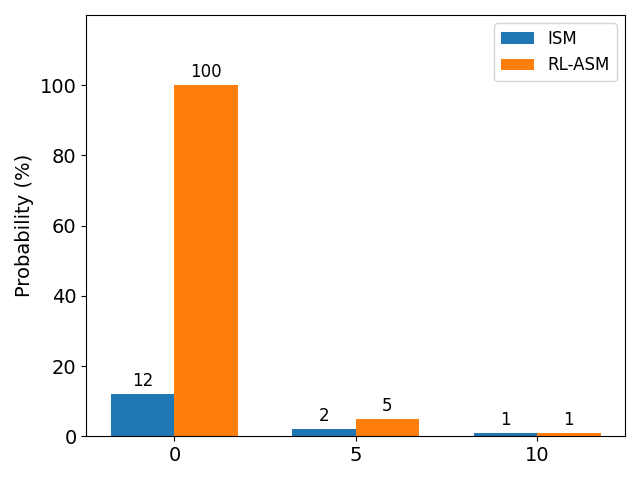} 
        \caption{EMAIL}
    \end{subfigure}
    \caption{The probabilities that ISM and our method find the optimal solutions within 600s.}
    \label{fig:effcience}
\end{figure*}

\subsubsection{Efficiency of Approximate Subgraph Matching}
Both ISM and RL-ASM utilize a branch-and-bound framework to effectively search the entire tree to find the optimal solution. Fig.~\ref{fig:effcience} illustrates the efficiencies of ISM and our method by reporting the probability of each method finding the optimal solution within 600 seconds. The results reveal that our method, guided by a neural network for action selection, is significantly more efficient than ISM. While both methods perform similarly with smaller graphs, such as the instances in the \textbf{AIDS} dataset, the efficiency gap is more pronounced with larger graphs. For example, on the \textbf{SYNTHETIC} dataset, our method significantly outperforms ISM, achieving a probability of identifying the optimal solution that is about 5x higher than that of ISM.

\begin{table}[t]
\centering
\caption{Ablation study of design choices on node-sorting and GraphGPS.}
\label{tab:ablation}\vspace{-5pt}
\begin{tabular}{@{}lccc@{}}
\toprule
Dataset          & \multicolumn{3}{c}{SYNTHETIC} \\ 
\cmidrule(l){2-4} 
Noise-Ratio       & 0\%      & 5\%      & 10\%     \\ 
\midrule
RL-ASM w/o node-sorting   & 0      & 1.63  & 3.38  \\
RL-ASM w/o GraphGPS    & 0      & 1.67  & 3.47  \\
RL-ASM           & 0      & 1.54   & 3.18   \\
\bottomrule
\end{tabular}
\end{table}

\subsubsection{Ablation Study}
In Section~\ref{sec:methods}, we introduce a method based on~\cite{bonnici2013subgraph} to sort the nodes in $G^q$ for mapping order $\phi$, and we also utilize GraphGPS~\cite{rampavsek2022recipe}, a type of graph transformer, as the intra-component in our encoder. This section assesses the impact of these design choices by replacing the sorted node order with a random node order or substituting GraphGPS with GatedGCN, which involves removing the global attention from the intra-component of our encoder. Table~\ref{tab:ablation} presents the results of these ablation studies. It is observed that both node sorting $\phi$ and GraphGPS enhance the performance of our RL-ASM. The node sorting prioritizes nodes with higher degrees and more matched neighbors, leveraging the structural information to facilitate accurate pattern matching. Moreover, GraphGPS, integrating an MPNN with global attention, significantly increases the expressiveness of our model.

\subsubsection{Generalization}
\begin{figure}[t]
\centering
\includegraphics[width=0.30\textwidth]{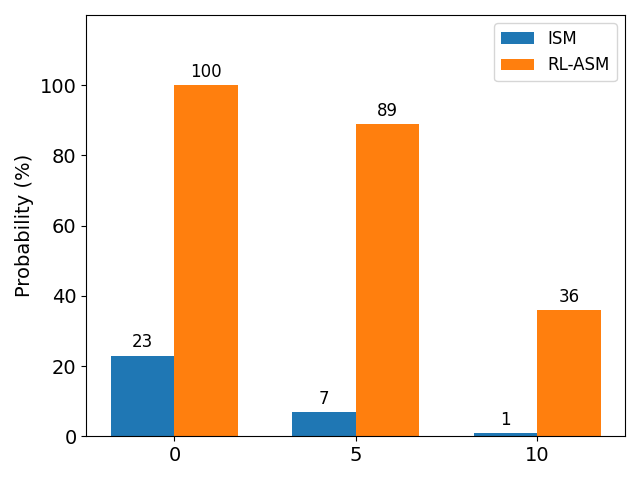}
  \caption{Efficiency on the random walk sampled dataset.}
  \label{fig:Rw}
\end{figure}

In the experiments above, we utilize the BFS to sample subgraphs and produce induced subgraphs (a.k.a query graphs) from the sampled nodes. In this section, we adopt a Random Walk (RW) approach for query graph sampling. Specifically, we start at a randomly selected node and move to an adjacent node with random walk, repeating this process for a pre-determined number of steps. During this process, we record all visited nodes and edges to construct the query graphs. Note that the RW sampled subgraphs may not necessarily be the induced subgraphs. Subsequently, we introduce noise to these sampled graphs to create noisy query graphs. This methodology is applied to the \textbf{SYNTHETIC} dataset to create training, validation, and test sets, which are used to assess the performance of ASM involving non-induced subgraphs. The GED between $G^q$ and the matched subgraphs, along with the probability of identifying the optimal solutions within 600 seconds, are reported in Table~\ref{tab:RW} and Fig.~\ref{fig:Rw}, respectively. The results indicate that our model still outperforms the baseline methods in terms of effectiveness and efficiency, even when the query graphs are not induced ones, demonstrating the generalization of RL-ASM.

\begin{table}[t]
\centering
\caption{Effectiveness on the random walk sampled dataset.}
\label{tab:RW}\vspace{-5pt}
\begin{tabular}{@{}lccc@{}}
\toprule
Dataset          & \multicolumn{3}{c}{SYNTHETIC} \\ 
\cmidrule(l){2-4} 
Noise-Ratio       & 0\%      & 5\%      & 10\%     \\ 
\midrule
ASP   & 0      & 2.03  & 4.04  \\
ISM    & 5.58    & 7.69  & 8.51  \\
RL-ASM  & 0      & 1.47  & 3.42   \\
\bottomrule
\end{tabular}
\label{tab:syn_rw}
\end{table}

\section{Conclusion}
As an NP-hard problem, approximate subgraph matching (ASM) is a fundamental research task in the database and graph mining communities. One significant strand of ASM methods follows the branch-and-bound search framework, where search performance is heavily influenced by the selection of actions (i.e., the mapped node pairs) in the search tree. We observe that existing methods, which utilize heuristics to select actions, lack the capability to fully exploit the structural and label information of graphs, and make sub-optimal decisions for ASM. In this paper, we introduce RL-ASM, a reinforcement learning based method for ASM that leverages Graph Transformer to extract the full graph information and optimizes an accumulative long-term rewards over episodes for ASM. Extensive experiments demonstrate the effectiveness and efficiency of our approach on four benchmark datasets. ASM is a relatively less investigated research area. We open source our code to facilitate the research in this area.

\bibliographystyle{IEEEtran}
\bibliography{bibfile.bib}

\end{document}